%% file: iclr2021_conference.tex
\documentclass{article} 
\input{math_commands.tex}

\UseRawInputEncoding
\usepackage{hyperref}
\usepackage{booktabs}
\usepackage{wrapfig}
\usepackage{url}
\usepackage{multirow}
\usepackage{amsthm}
\usepackage{graphicx}
\usepackage[draft]{changes}

\theoremstyle{definition}
\newtheorem{definition}{Definition}[section]

\usepackage{algorithm}
\usepackage{algorithmicx}
\usepackage{xspace}

\newcommand{\ibsubgraph}{IB-subgraph\xspace}

\usepackage{algpseudocode}
\usepackage{geometry}
\usepackage[T1]{fontenc}
\usepackage[utf8]{inputenc}
\usepackage{authblk}

\definechangesauthor[color=blue, name="Tingyang Xu"]{TX}
\definechangesauthor[color=orange, name="Junchi Yu"]{JY}
\definechangesauthor[name="Yatao Bian"]{YB}
\definechangesauthor[name="Roy Rong"]{RR}

\title{Graph Information Bottleneck for Subgraph Recognition}

\author[1,2,3]{Junchi Yu}
\author[3]{Tingyang Xu}
\author[3]{Yu Rong}
\author[3]{Yatao Bian}
\author[3]{Junzhou Huang}
\author[1,2,4]{Ran He}
\affil[1]{NLPR\&CRIPAC, Institute of Automation, Chinese Academy of Sciences}
\affil[2]{University of Chinese Academy of Sciences}
\affil[3]{Tencent AI LAB}
\affil[4]{Center for Excellence in Brain Science and Intelligence Technology, CAS}

\begin{document}

\maketitle

\begin{abstract}
Given the input graph and its label/property, several key problems  of graph learning, such as finding interpretable subgraphs, graph denoising and graph compression,  can be  attributed to the fundamental problem of recognizing a subgraph of the original one.  This subgraph shall be as informative as possible, yet contains less redundant and noisy structure. This problem setting is closely related to the well-known information bottleneck (IB) principle, which, however, has less been studied for the irregular graph data and graph neural networks (GNNs). In this paper, we propose a framework of Graph Information Bottleneck (GIB) for the subgraph recognition problem in deep graph learning. Under this framework, one can recognize the maximally informative yet compressive subgraph, named \ibsubgraph.  However, the GIB objective is notoriously hard to optimize, mostly due to the intractability of the mutual information of irregular graph data and the unstable optimization process. In order to tackle these challenges, we propose: 
i) a GIB objective based-on a mutual information estimator for the irregular graph data; 
ii) a bi-level optimization scheme to maximize the GIB objective; 
iii) a connectivity loss to stabilize the optimization process. We evaluate the properties of the \ibsubgraph in three application scenarios: improvement of graph classification, graph interpretation and graph denoising. Extensive experiments demonstrate that the information-theoretic  \ibsubgraph  enjoys superior graph properties. 
\end{abstract}

\section{Introduction}

Classifying the underlying labels or properties of graphs is a fundamental problem in deep graph learning with applications across many fields, 
such as biochemistry and social network analysis. 
However, real world graphs are likely to contain redundant even noisy information \cite{conf/icml/FranceschiNPH19, journals/corr/abs-1911-07123}, which poses a huge negative impact for graph classification. This triggers an interesting problem of recognizing an informative yet compressed subgraph from the original graph. 
For example, in drug discovery, when viewing molecules as graphs with atoms as nodes and chemical bonds as edges, biochemists are interested in identifying the subgraphs that mostly represent certain properties of the molecules, namely the functional groups \cite{substructure,mmpn}. In graph representation learning, the predictive subgraph highlights the vital substructure for graph classification, and provides an alternative way for yielding graph representation besides mean/sum aggregation \cite{GCN,velickovic2017graph,Xu:2019ty} and pooling aggregation \cite{diffpool,sapool,specpool}. In graph attack and defense, it is vital to purify a perturbed graph and mine the robust structures for classification \cite{journals/corr/abs-2005-10203}.

Recently, the mechanism of self-attentive aggregation \cite{hierG} somehow discovers a vital substructure at node level with a well-selected threshold. However, this method only identifies isolated important nodes but ignores the topological information at subgraph level. 
Consequently, it leads to a novel challenge as subgraph recognition: 
\emph{How can we recognize a compressed subgraph with minimum information loss in terms of predicting  the graph labels/properties?}

Recalling the above challenge, there is a similar problem setting in information theory called information bottleneck (IB) principle \cite{ib}, which aims to juice out a compressed data from the original data that keeps most predictive information of labels or properties. Enhanced with deep learning, IB can learn informative representation from regular data in the fields of computer vision \cite{rl_ib1,vib,conf/iccv/LuoLGYY19}, reinforcement learning \cite{rl_ib2,rl_ib3} and natural language precessing \cite{nlp_ib1}. However, current IB methods, like VIB \cite{vib}, is still incapable for irregular graph data. It is still challenging for IB to compress irregular graph data, like a subgraph from an original graph, with a minimum information loss.

Hence, we advance the IB principle for irregular graph data to resolve the proposed subgraph recognition problem, which leads to a novel principle, Graph Information Bottleneck (GIB). Different from prior researches in IB that aims to learn an optimal representation of the input data in the hidden space, GIB directly reveals the vital substructure in the subgraph level. 
We first i) leverage the mutual information estimator from Deep Variational Information Bottleneck (VIB) \cite{vib} for irregular graph data as the GIB objective. 
However, VIB is intractable to compute the mutual information without knowing the distribution forms, especially on graph data. 
To tackle this issue, ii) we adopt a bi-level optimization scheme to maximize the GIB objective. 
Meanwhile, the continuous relaxation that we adopt to approach the discrete selection of subgraph will lead to unstable optimization process. 
To further stabilize the training process and encourage a compact subgraph, iii) we propose a novel connectivity loss to assist GIB to effectively discover the maximally informative but compressed subgraph, which is defined as \emph{\ibsubgraph}. 
By optimizing the above GIB objective and connectivity loss, one can recognize the \ibsubgraph without any explicit subgraph annotation. 
On the other hand, iv) GIB is model-agnostic and can be easily plugged into various Graph Neural Networks (GNNs).

We evaluate the properties of the \ibsubgraph 
in three application scenarios:  improvement of graph classification, graph interpretation, and graph denoising.  Extensive experiments on both synthetic and real world datasets demonstrate that the information-theoretic \ibsubgraph enjoys superior graph properties compared to the subgraphs found by SOTA baselines.

\section{Related Work}

\textbf{Graph Classification.} In recent literature, there is a surge of interest in adopting graph neural networks (GNN) in graph classification. The core idea  is to aggregate all the  node information for graph representation. A typical implementation  is the mean/sum aggregation \cite{GCN,Xu:2019ty}, which is to average or sum up the node embeddings. An alternative way is to leverage the hierarchical structure of graphs, which leads to the pooling aggregation \cite{diffpool,sortpool,sapool,specpool}. When tackling with the redundant and noisy graphs, these approaches will likely to result in sub-optimal graph representation.

\textbf{Information Bottleneck.} Information bottleneck (IB), originally proposed for signal processing, attempts to find a short code of the input signal but preserve maximum information of the code \cite{ib}. \cite{vib} firstly bridges the gap between IB and the deep learning, and proposed variational information bottleneck (VIB). Nowadays, IB and VIB have been wildly employed in computer vision \cite{rl_ib1,conf/iccv/LuoLGYY19}, reinforcement learning \cite{rl_ib2,rl_ib3}, natural language processing \cite{nlp_ib1} and speech and acoustics \cite{speech_ib1} due to the capability of learning compact and meaningful representations. However, IB is less researched on irregular graphs due to the intractability of mutual information.

\textbf{Subgraph Discovery.} Traditional subgraph discovery includes dense subgraph discovery and frequent subgraph mining. Dense subgraph discovery aims to find the subgraph with the highest density (e.g. the number of edges over the number of nodes \cite{densesub,conf/kdd/GionisT15}). Frequent subgraph mining is to look for the most common substructure among graphs \cite{gspan,subdue,SLEUTH}. Recently, researchers discover the vital substructure at node level via the attention mechanism \cite{velickovic2017graph,sapool,conf/nips/KnyazevTA19}. \cite{gnnexplainer} further identifies the important computational graph for node classification. \cite{sgnn} discovers subgraph representations with specific topology given subgraph-level annotation.

\section{Notations and Preliminaries}

Let $\{(\gG_1, Y_1),\dots,(\gG_N, Y_N)\}$ be a set of $N$ graphs with their real value properties or categories, where $\gG_n$ refers to the $n$-th graph and $Y_n$ refers to the corresponding properties or labels. We denote by $\gG_n=(\sV,\mathbb{E}, \mA, \mX)$ the $n$-th graph of size $\mM_n$ with node set $\sV=\{V_i|i=1,\dots,\mM_n\}$, edge set $\mathbb{E}=\{(V_i, V_j)|i>j; V_i,V_j \text{ is connected}\}$,
adjacent matrix $\mA\in \{0,1\}^{\mM_n\times \mM_n}$, and feature matrix $\mX\in \sR^{\mM_n\times d}$ of $\mV$ with $d$ dimensions, respectively. Denote the neighborhood of $V_i$ as $\mathcal{N}(V_i)=\{V_j|(V_i, V_j)\in \mathbb{E}\}$. 
We use 
$\gG_{sub}$ as a specific subgraph and $\overline{\gG}_{sub}$ as the complementary structure of $\gG_{sub}$ in $\gG$. Let $f:\sG \rightarrow \sR / [0,1,\cdots,n] $ be the mapping from graphs to the real value property or category, $\mY$, $\sG$ is the domain of the input graphs. $I(\mX,\mY)$ refers to the Shannon mutual information of two random variables.

\subsection{Graph convolutional network}

Graph convolutional network (GCN) is widely adopted to graph classification.  Given a graph $\gG = (\sV,\mathbb{E})$ with node feature $\mX$ and adjacent matrix $\mA$, GCN outputs the node embeddings $\mX^{'}$ from the following process:
\begin{equation}
\begin{aligned}
\mX^{'} = \mathrm{GCN}(\mA,\mX;\mW) = \mathrm{ReLU}(\mD^{-\frac{1}{2}}\hat\mA\mD^{-\frac{1}{2}}\mW), 
\end{aligned}
\end{equation}
where $\mD$ refers to the diagonal matrix with nodes' degrees and $\mW$ refers to the model parameters.

One can simply sum up the node embeddings to get a fixed length graph embeddings \cite{Xu:2019ty}. Recently, researchers attempt to exploit hierarchical structure of graphs, which leads to various graph pooling methods \cite{hierG,conf/icml/GaoJ19,sapool,journals/corr/abs-1905-10990,sortpool,asapool,diffpool}. \cite{hierG} enhances the graph pooling with self-attention mechanism to leverage the importance of different nodes contributing to the results. Finally, the graph embedding is obtained by multiplying the node embeddings with the normalized attention scores:
\begin{equation}
\begin{aligned}
\mE = \mathrm{Att}(\mX^{'}) =\mathrm{ softmax}(\Phi_{2}\mathrm{tanh}(\Phi_{1}\mX^{'T}))\mX^{'},
\end{aligned}
\end{equation}
where $\Phi_1$ and $\Phi_2$ refers to the model parameters of self-attention.
\subsection{Optimizing Information bottleneck objective}
Given the input signal $X$ and the label $Y$, the objective of IB is maximized to find the  the internal code $Z$:  $\max_{Z}{I(Z,Y)-\beta I(X,Z)}$, 
where $\beta$ refers to a hyper-parameter trading off informativeness and compression. 
Optimizing this objective will lead to a compact but informative $Z$. \cite{vib} optimize a tractable lower bound of the IB objective: 
\begin{equation}
\begin{aligned}
\mathcal{L}_{VIB} = \frac{1}{N} \sum\nolimits_{i=1}^{N} \int\nolimits p(z|x_{i})\log{q_{\phi}(y_{i}|z)} dz  - \beta \mathrm{KL}(p(z|x_{i})|r(z)),
\label{vib_obj}
\end{aligned}
\end{equation}
where $q_{\phi}(y_{i}|z)$ is the variational approximation to $p_{\phi}(y_{i}|z)$ and $r(z)$ is the prior distribution of $Z$. 
However, it is hard to estimate the mutual information in high dimensional space when the distribution forms are inaccessible, especially for irregular graph data.

\begin{figure}[t]
\vspace{-0.9cm}
\begin{center}
\centerline{\includegraphics[width=0.8\columnwidth]{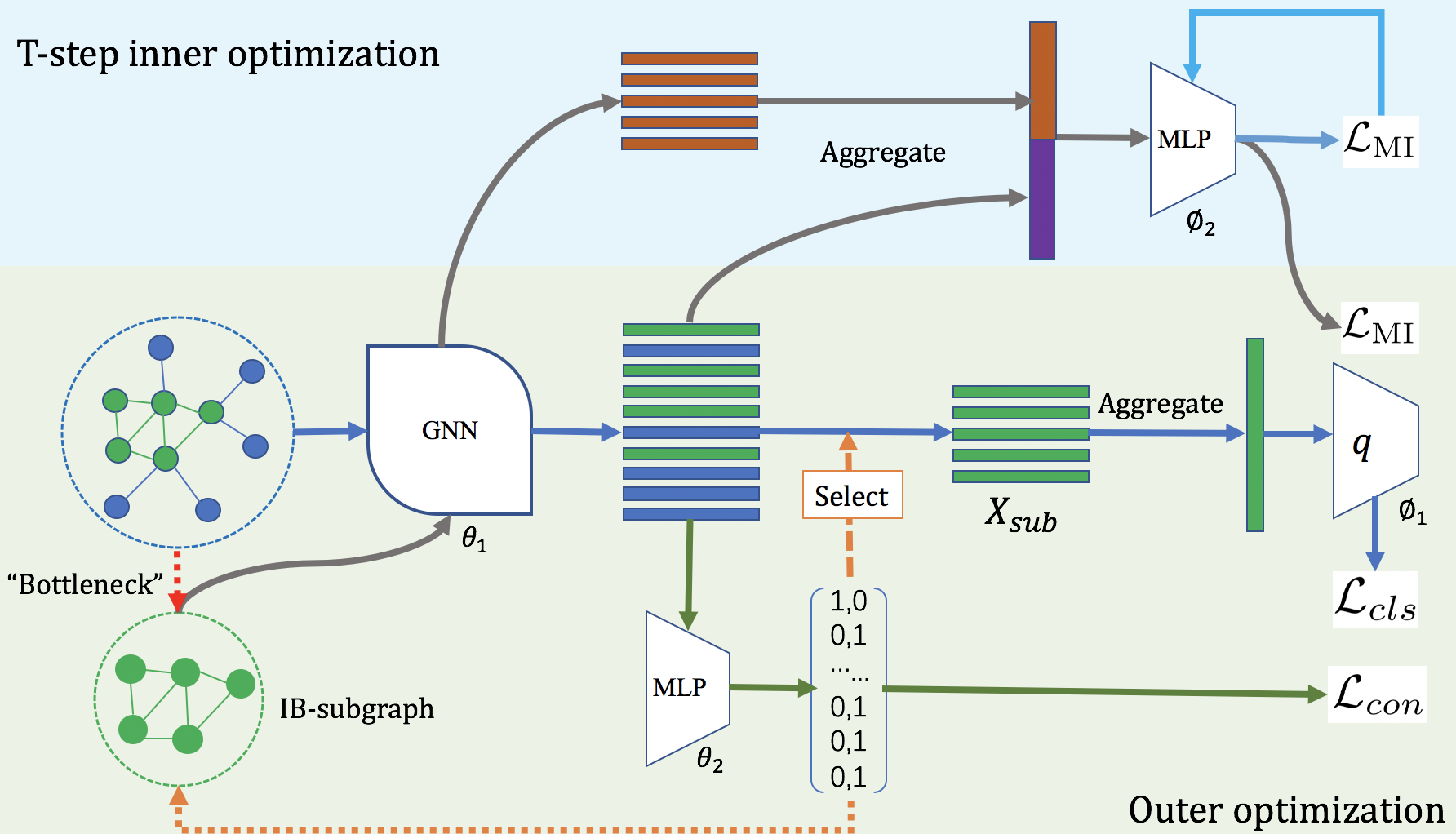}}
\end{center}
\vspace{-0.6cm}
\caption{Illustration of the  proposed graph information  bottleneck (GIB) framework. We employ a bi-level optimization scheme to optimize the GIB objective and thus yielding the \ibsubgraph. In the  inner optimization phase, we estimate $I(\gG,\gG_{sub})$ by optimizing the statistics network of the DONSKER-VARADHAN representation \cite{dv-representation}.  Given a good estimation of $I(\gG,\gG_{sub})$, in the outer optimization phase, we maximize the GIB objective by optimizing the mutual information,  the classification loss $\mathcal{L}_{cls}$ and connectivity loss $\mathcal{L}_{con}$. } 
\label{flowchart}
\end{figure}

\section{Optimizing the Graph Information Bottleneck Objective for Subgraph Recognition}

In this section, we will elaborate the proposed method in details. We first formally define the graph information bottleneck and \ibsubgraph. Then, we introduce a novel framework for GIB  to effectively find the \ibsubgraph. We further propose a bi-level optimization scheme and a graph mutual information estimator for GIB optimization. Moreover, we do a continuous relaxation to the generation of subgraph, and propose a novel loss to stabilize the training process.  

\subsection{graph information bottleneck}

We generalize the information bottleneck principle to learn a informative representation of irregular graphs, which leads to the graph information bottleneck (GIB) principle.

\theoremstyle{definition}
\begin{definition}[Graph Information Bottleneck]
Given a graph $\gG$ and its label $Y$, the GIB seeks for the most informative yet compressed representation $Z$ by optimizing the following objective:
\begin{equation}
\begin{aligned}
&\max_{Z}{I(Y,Z)}  \text{ s.t. } I(\gG,Z) \leq I_{c}.
\end{aligned}
\label{gib}
\end{equation}
where $I_{c}$ is the information constraint between $\gG$ and $Z$. By introducing a Lagrange multiplier $\beta$ to Eq.~\ref{gib}, we reach its  unconstrained form:
\begin{equation}
\begin{aligned}
\max_{Z}{ I(Y,Z) - \beta I(\gG,Z) }.
\end{aligned}
\label{gib-dual}
\end{equation}
\end{definition}
Eq.~\ref{gib-dual} gives a general formulation of GIB. Here, in subgraph recognition, we focus on a subgraph which is compressed with minimum information loss in terms of graph properties.

\begin{definition}[\ibsubgraph]
For a graph $\gG$, its maximally informative yet compressed subgraph, namely \ibsubgraph can be obtained by optimizing the following objective:
%
\begin{equation}
\begin{aligned}
&\max_{\gG_{sub}\in \sG_{sub}} I(Y,\gG_{sub})-\beta I(\gG,\gG_{sub}).
\end{aligned}
\label{gib-sub}
\end{equation}
where $\sG_{sub}$ indicates the set of all subgraphs of $\gG$.

\end{definition}

\ibsubgraph enjoys various pleasant properties and can be applied to multiple graph learning tasks such as improvement of graph classification, graph interpretation, and graph denoising. However, the GIB objective in Eq.~\ref{gib-sub} is notoriously hard to optimize due to the intractability of mutual information and the discrete nature of irregular graph data. We then introduce approaches on  how to optimize such objective and derive the \ibsubgraph.

\subsection{Bi-level optimization for the GIB objective}

The GIB objective in Eq.~\ref{gib-sub} consists of two parts. We examine the first term $I(Y,\gG_{sub})$ in Eq.~\ref{gib-sub}, first. This term measures the relevance between $\gG_{sub}$ and $Y$. We expand $I(Y,\gG_{sub})$ as:  
\begin{equation}
\begin{aligned}
I(Y,\gG_{sub}) = \int p(y,\gG_{sub}) \log{{p(y|\gG_{sub})}} dy \  d\gG_{sub} + \mathrm{H}(Y). \label{sub-obj}\\
\end{aligned}
\end{equation}
$H(Y)$ is the entropy of $Y$ and thus can be ignored. In practice, we approximate $p(y,\gG_{sub})$ with an empirical distribution $p(y,\gG_{sub}) \approx\frac{1}{N} \sum_{i=1}^{N}\delta_{y_i}(y)\delta_{\gG_{sub,i}}(\gG_{sub})$, where $\gG_{sub}$ is the output subgraph and $Y$ is the graph label. By substituting the true posterior $p(y|\gG_{sub})$ with a variational approximation $q_{\phi_{1}}(y|\gG_{sub})$, we obtain a tractable lower bound of the first term in Eq.~\ref{gib-sub}:
\begin{equation}
\begin{aligned}
I(Y,\gG_{sub}) &\geq \int p(y,\gG_{sub}) \log{{q_{\phi_{1}}(y|\gG_{sub})}} dy \ d\gG_{sub} \\
&\approx \frac{1}{N} \sum_{i=1}^{N} q_{\phi_{1}}(y_{i}|\gG_{sub_{i}}) =: -\mathcal{L}_{cls}(q_{\phi_{1}}(y|\gG_{sub}),y_{gt}), \\
\end{aligned}
\label{mi_1}
\end{equation}
where $y_{gt}$ is the ground truth label of the graph. Eq.~\ref{mi_1} indicates that maximizing $I(Y,\gG_{sub})$ is achieved by the minimization of the classification loss between $Y$ and $\gG_{sub}$ as  $\mathcal{L}_{cls}$. Intuitively, minimizing $\mathcal{L}_{cls}$ encourages the subgraph to be predictive of the graph label. In practice, we choose the cross entropy loss for categorical $Y$ and the mean squared loss for continuous $Y$, respectively. For more details of deriving Eq.~\ref{sub-obj} and Eq.~\ref{mi_1}, please refer to Appendix \ref{appendix_eq9}.

Then, we consider the minimization of $I(\gG,\gG_{sub})$ which is the second term of Eq.~\ref{gib-sub}. Remind that \cite{vib} introduces a tractable prior distribution $r(Z)$ in Eq.~\ref{vib_obj}, and thus results in a variational upper bound. However, this setting is troublesome as it is hard to find a reasonable prior distribution for $p(\gG_{sub})$, which is the distribution of graph substructures instead of latent representation. Thus we go for another route.  Directly applying the DONSKER-VARADHAN representation \cite{dv-representation} of the KL-divergence, we have:
\begin{equation}
\begin{aligned}
I(\gG,\gG_{sub}) = \sup \limits_{f_{\phi_2}:\sG\times \sG\rightarrow \sR} \mathbb{E}_{\gG,\gG_{sub}\in p(\gG,\gG_{sub})}f_{\phi_{2}}(\gG,\gG_{sub})-\log{\mathbb{E}_{\gG \in p(\gG),\gG_{sub}\in p(\gG_{sub})}e^{f_{\phi_{2}}(\gG,\gG_{sub})}},
\end{aligned}
\label{mi_est}
\end{equation}
where $f_{\phi_2}$ is the statistics network that maps from the graph set to the set of real numbers. 
In order to approximate $I(\gG,\gG_{sub})$ using Eq.~\ref{mi_est}, we design a statistics network  based on modern GNN architectures as shown by Figure \ref{flowchart}: first we use a GNN to extract  embeddings from both $\gG$ and $\gG_{sub}$ (parameter shared with the subgraph generator, which will be elaborated in Section \ref{sec_subgraph_generator}), then concatenate $\gG$ and $\gG_{sub}$ embeddings and feed them into a MLP, which finally produces the real number.    
In conjunction with the sampling method to approximate $p(\gG,\gG_{sub})$, $p(\gG)$ and $p(\gG_{sub})$, we reach the following optimization problem to approximate\footnote{Notice that the MINE estimator \cite{mine} straightforwardly uses the DONSKER-VARADHAN representation to derive an MI estimator between the  regular input data and its vectorized representation/encoding. It cannot be applied to estimate the mutual information between $\gG$ and $\gG_{sub}$ since both of $\gG$ and $\gG_{sub}$ are irregular graph data.} $I(\gG,\gG_{sub})$: 
\begin{equation}
\begin{aligned}
\max_{\phi_{2}} \quad \mathcal{L}_{\mathrm{MI}}(\phi_{2},\gG_{sub}) =  \frac{1}{N}\sum_{i=1}^{N}f_{\phi_{2}}(\gG_{i},\gG_{sub,i})-\log{\frac{1}{N}\sum_{i=1,j\neq i}^{N}e^{f_{\phi_{2}}(\gG_{i},\gG_{sub,j})}}.
\end{aligned}
\label{mi_loss}
\end{equation}

With the approximation to the MI in graph data, we combine Eq.~\ref{gib-sub} , Eq.~\ref{mi_1} and Eq.~\ref{mi_loss} and formulate the optimization process of GIB as a tractable bi-level optimization problem:
\begin{align}
&\min \limits_{\gG_{sub},\phi_{1}} \quad \mathcal{L}(\gG_{sub},\phi_{1},\phi_{2}^{*}) = \mathcal{L}_{cls}(q_{\phi_{1}}(y|\gG_{sub}),y_{gt}) + \beta \mathcal{L}_{\rm MI}(\phi_{2}^{*},\gG_{sub}) \label{bilevel}\\
&\text{ s.t. } \quad \phi_{2}^{*} = \mathop{\argmax} \limits_{\phi_{2}}\mathcal{L}_{\mathrm{MI}}(\phi_{2},\gG_{sub}) \label{t-steps}.
\end{align}
We first derive a sub-optimal $\phi_2$ notated as $\phi_2^*$ by optimizing Eq.~\ref{t-steps} for T steps as inner loops. After the T-step optimization of the inner-loop ends, Eq.~\ref{mi_loss} is a proxy for MI minimization for the GIB objective as an outer loop. Then, the parameter $\phi_{1}$ and the subgraph $\gG_{sub}$ are optimized to yield \ibsubgraph.
However, in the outer loop, the discrete nature of $\gG$ and $\gG_{sub}$ hinders applying the gradient-based method to optimize the bi-level objective and find the \ibsubgraph.   

\subsection{The Subgraph Generator and  connectivity loss}
\label{sec_subgraph_generator}

To alleviate the discreteness in Eq.~\ref{bilevel}, we propose the continuous relaxation to the subgraph recognition and propose a loss to stabilize the training process.

\textbf{Subgraph generator:} For the input graph $\gG$, we generate its \ibsubgraph with the node assignment $\mS$ which indicates the node is in $\gG_{sub}$ or $\overline\gG_{sub}$. Then, we introduce a continuous relaxation to the node assignment with the probability of nodes belonging to the $\gG_{sub}$ or $\overline\gG_{sub}$. For example, the $i$-th row of $\mS$ is 2-dimensional vector $\textbf{[}p(V_{i}\in \gG_{sub}|V_{i}), p(V_{i} \in \overline\gG_{sub}|V_{i})\textbf{]}$. We first use an $l$-layer GNN to obtain the node embedding and employ a multi-layer perceptron (MLP) to output $\mS$ :
\begin{equation}
\begin{aligned}
\mX^{l}& = \mathrm{GNN}(\mA,\mX^{l-1};\theta_{1}), \quad 
\mS &= \mathrm{MLP}(\mX^{l};\theta_{2}).
\end{aligned}
\end{equation}
$\mS$ is a $n\times 2$ matrix, where $n$ is the number of nodes. For simplicity, we compile the above modules as the subgraph generator, denoted as $g(;\theta)$ with $\theta:=(\theta_1, \theta_2)$. When $\mS$ is well-learned, the assignment of nodes is supposed to saturate to 0/1. The representation of $\gG_{sub}$, which is employed for predicting the graph label, can be obtained by taking the first row of $\mS^{T}\mX^{l}$.

\textbf{Connectivity loss:} However, poor initialization will cause $p(V_{i}\in \gG_{sub}|V_{i})$ and $p(V_{i}\in \overline\gG_{sub}|V_{i})$ to be close. This will either lead the model to assign all nodes to $\gG_{sub}$ / $\overline\gG_{sub}$, or result that the representations of $\gG_{sub}$ contain much information from the redundant nodes. These two scenarios will cause the training process to be unstable. On the other hand, we suppose our model to have an inductive bias to better leverage the topological information while $\mS$ outputs the subgraph at a node-level.    Therefore, we propose the following connectivity loss:  
\begin{equation}
\begin{aligned}
\mathcal{L}_{con} = || \mathrm{Norm}(\mS^{T}\mA\mS)- \mI_2||_F,
\end{aligned}
\label{L_con}
\end{equation}
where $\mathrm{Norm(\cdot)}$ is the row-wise normalization, $||\cdot||_F$ is the Frobenius norm, and $\mI_2$ is a $2\times2$ identity matrix. %
$\mathcal{L}_{con}$ not only leads to distinguishable node assignment, but also encourage the subgraph to be compact.  Take $(\mS^{T}\mA\mS)_{1:}$ for example, denote $a_{11},a_{12}$ the element 1,1 and the element 1,2 of $\mS^{T}\mA\mS$,
\begin{equation}
\begin{aligned}
a_{11} = \sum_{i,j} A_{ij}p(V_{i}\in \gG_{sub}|V_{i})p(V_{j}\in \gG_{sub}|V_{j}),\\
a_{12} = \sum_{i,j} A_{ij}p(V_{i}\in \gG_{sub}|V_{i})p(V_{j}\in \overline \gG_{sub}|V_{j}).\\
\end{aligned}
\end{equation}
Minimizing $\mathcal{L}_{con}$ results in $\frac{a_{11}}{a_{11}+a_{12}}\rightarrow 1$. This occurs if $V_{i}$ is in $ \gG_{sub}$, the elements of $\mathcal{N}(V_{i})$ have a high probability in $ \gG_{sub}$.  Minimizing $\mathcal{L}_{con}$ also encourages $\frac{a_{12}}{a_{11}+a_{12}}\rightarrow 0$. This encourages $p(V_{i}\in \gG_{sub}|V_{i})\rightarrow 0/1$ and less cuts between $\gG_{sub}$ and $\overline \gG_{sub}$. This also holds for $\overline \gG_{sub}$ when analyzing $a_{21}$ and $a_{22}$.

In a word, $\mathcal{L}_{con}$ encourages distinctive $\mS$ to stabilize the training process and a compact topology in the subgraph. Therefore, the overall loss is:
\begin{equation}
\begin{aligned}
&\min \limits_{\theta,\phi_{1}} \quad \mathcal{L}(\theta,\phi_{1},\phi_{2}^{*}) = \mathcal{L}_{con}(g(\gG;\theta)) + \mathcal{L}_{cls}(q_{\phi_{1}}(g(\gG;\theta)),y_{gt}) + \beta \mathcal{L}_{\rm MI}(\phi_{2}^{*},\gG_{sub}) \\
&\text{ s.t. }\quad \phi_{2}^{*} = \mathop{\argmax} \limits_{\phi_{2}}\mathcal{L}_{\rm MI}(\phi_{2},\gG_{sub}).
\end{aligned}
\label{final-bilevel}
\end{equation}
We provide the pseudo code in the Appendix to better illustrate how to optimize the above objective.

\begin{table}[t]
  \centering
  \caption{Classification accuracy in percent. The pooling methods yield pooling aggregation while the backbones yield mean aggregation. The proposed GIB method with backbones yields subgraph embedding by aggregating the nodes in subgraphs.}
    \begin{tabular}{ccccc}
    \toprule
    \textbf{Method} & \textbf{MUTAG} & \textbf{PROTEINS} & \textbf{IMDB-BINARY} & \textbf{DD} \\
    \midrule
     SortPool & \textbf{0.844 $\pm$ 0.141} & 0.747 $\pm$ 0.044 & 0.712 $\pm$ 0.047 & 0.732 $\pm$ 0.087 \\
           ASAPool  & 0.743 $\pm$ 0.077 & 0.721 $\pm$ 0.043 & 0.715 $\pm$ 0.044 & 0.717 $\pm$ 0.037 \\
           DiffPool & 0.839 $\pm$ 0.097 & 0.727 $\pm$ 0.046 & 0.709 $\pm$ 0.053 &  0.778 $\pm$ 0.030 \\
           EdgePool & 0.759 $\pm$ 0.077 & 0.723 $\pm$ 0.044 & 0.728 $\pm$ 0.044 & 0.736 $\pm$ 0.040 \\
           AttPool & 0.721 $\pm$ 0.086 & 0.728 $\pm$ 0.041 & 0.722 $\pm$ 0.047 & 0.711 $\pm$ 0.055 \\
    \midrule
     GCN   & 0.743$\pm$0.110 & 0.719$\pm$0.041 & 0.707 $\pm$ 0.037 & 0.725 $\pm$ 0.046 \\
           GraphSAGE & 0.743$\pm$0.077 & 0.721 $\pm$ 0.042 & 0.709 $\pm$ 0.041 & 0.729 $\pm$ 0.041 \\
           GIN   & 0.825$\pm$0.068 & 0.707 $\pm$ 0.056 & 0.732 $\pm$ 0.048 & 0.730 $\pm$ 0.033 \\
           GAT   & 0.738 $\pm$ 0.074 & 0.714 $\pm$ 0.040 & 0.713 $\pm$ 0.042 & 0.695 $\pm$ 0.045 \\
    \midrule
     \textbf{GCN+GIB} & 0.776 $\pm$ 0.075 & 0.748 $\pm$ 0.046 & 0.722 $\pm$ 0.039 & 0.765 $\pm$  0.050 \\
     \textbf{GraphSAGE+GIB} & 0.760  $\pm$ 0.074 & 0.734 $\pm$ 0.043 & 0.719 $\pm$  0.052 & \textbf{0.781 $\pm$ 0.042} \\
      \textbf{GIN+GIB} & 0.839 $\pm$  0.064 & \textbf{0.749 $\pm$  0.051} & \textbf{0.737 $\pm$ 0.070} & 0.747 $\pm$  0.039 \\
      \textbf{GAT+GIB} & 0.749 $\pm$ 0.097 & 0.737 $\pm$ 0.044 & 0.729 $\pm$  0.046 & 0.769 $\pm$  0.040 \\
    \bottomrule
    \end{tabular}%
  \label{tab:2}%
\end{table}%

\begin{table}[t]
  \centering
  \caption{The mean and variance of absolute property bias between the graphs and the corresponding subgraphs. Note that we try several initiations for GCN+GIB w/o $\mathcal{L}_{con}$ and $\mathcal{L}_{MI}$ to get the current results due to the instability of optimization process.}
    \begin{tabular}{ccccc}
    \toprule
    \textbf{Method}    & \textbf{QED} & \textbf{DRD2}  & \textbf{HLM-CLint}   & \textbf{MLM-CLint} \\
    \midrule
    GCN+Att05 & 0.48$\pm$ 0.07 & 0.20$\pm$ 0.13 & 0.90$\pm$ 0.89 & 0.92$\pm$ 0.61 \\
    GCN+Att07 & 0.41$\pm$ 0.07 & 0.16$\pm$ 0.11 & 1.18$\pm$ 0.60 & 1.69$\pm$ 0.88 \\
    \midrule
    GCN+GIB w/o $\mathcal{L}_{con}$ & 0.46$\pm$ 0.07 & 0.15$\pm$ 0.12 & 0.45$\pm$ 0.37 & 1.58$\pm$ 0.86 \\
    GCN+GIB w/o $\mathcal{L}_{MI}$ & 0.43$\pm$ 0.15 & 0.21$\pm$ 0.13 & 0.48$\pm$ 0.34 & 1.20$\pm$ 0.97 \\
    \textbf{GCN+GIB}  & \textbf{0.38$\pm$ 0.12} & \textbf{0.06$\pm$ 0.09} & \textbf{0.37$\pm$ 0.30} & \textbf{0.72$\pm$ 0.55} \\
    \bottomrule
    \end{tabular}%
  \label{tab:1}%
\end{table}%

\section{Experiments}
In this section, we evaluate the proposed GIB method on three scenarios, including improvement of graph classification, graph interpretation and graph denoising.

\subsection{Baselines and settings}

\textbf{Improvement of graph classification:} For improvement of graph classification, GIB generates graph representation by aggregating the subgraph information. We plug GIB into various backbones including GCN \cite{GCN}, GAT \cite{velickovic2017graph}, GIN \cite{Xu:2019ty} and GraphSAGE \cite{conf/nips/HamiltonYL17}. We compare the proposed method with the mean/sum aggregation \cite{GCN,velickovic2017graph,conf/nips/HamiltonYL17,Xu:2019ty} and pooling aggregation \cite{sortpool,asapool,diffpool,journals/corr/abs-1905-10990} in terms of classification accuracy. 

\textbf{Graph interpretation:} The goal of graph interpretation is to find the substructure which shares the most similar property to the molecule. If the substructure is disconnected, we evaluate its largest connected part. We compare GIB with the attention mechanism \cite{hierG}. That is, we attentively aggregate the node information for graph prediction. The interpretable subgraph is generated by choosing the nodes with top $50\%$ and $70\%$ attention scores, namely Att05 and Att07. GIB outputs the interpretation with the \ibsubgraph. Then, we evaluate the absolute property bias (the absolute value of the difference between the property of graph and subgraph) between the graph and its interpretation. For fare comparisons, we adopt the same GCN as the backbone for different methods.

\textbf{Graph denoising:} We translate the permuted graph into the line-graph and use GIB and attention to 1) infer the real structure of graph, 2) classify the permuted graph via the inferred structure. We further compare the performance of GCN and DiffPool on the permuted graphs.

\subsection{datasets}

\textbf{Improvement of graph classification:} We evaluate different methods on the datasets of  \textbf{MUTAG} \cite{PhysRevLett.108.058301}, \textbf{PROETINS} \cite{conf/ismb/BorgwardtOSVSK05}, \textbf{IMDB-BINARY} and \textbf{DD} \cite{nr} datasets.  \footnote{We follow the protocol in https://github.com/rusty1s/pytorch\_geometric/tree/master/benchmark/kernel}. 

\textbf{Graph interpretation:} We construct the datasets for graph interpretation on four molecule properties based on ZINC dataset, which contains 250K molecules. \textbf{QED} measures the drug likeness of a molecule, which is bounded within the range $(0,1.0)$. \textbf{DRD2} measures the probability that a molecule is active against dopamine type 2 receptor, which is bounded with $(0,1.0)$. \textbf{HLM-CLint} and \textbf{MLM-CLint} are  estimate values of in vitro human and mouse liver microsome metabolic stability (base 10 logrithm of mL/min/g). We sample the molecules with QED $ \geq 0.85$, DRD2 $ \geq 0.50$, HLM-CLint $\geq 2$,  MLM-CLint $\geq 2$ for each task. We use $85\%$ of these molecules for training, $5\%$ for validating and $10\%$ for testing.  
 
\textbf{Graph denoising:} We generate a synthetic dataset by adding $30\%$ redundant edges for each graph in \textbf{MUTAG} dataset. We use $70\%$ of these graphs for training, $5\%$ for validating and $25\%$ for testing. 

\begin{figure}[h]
\vspace{-1cm}
\begin{center}
\centerline{\includegraphics[width=1\columnwidth]{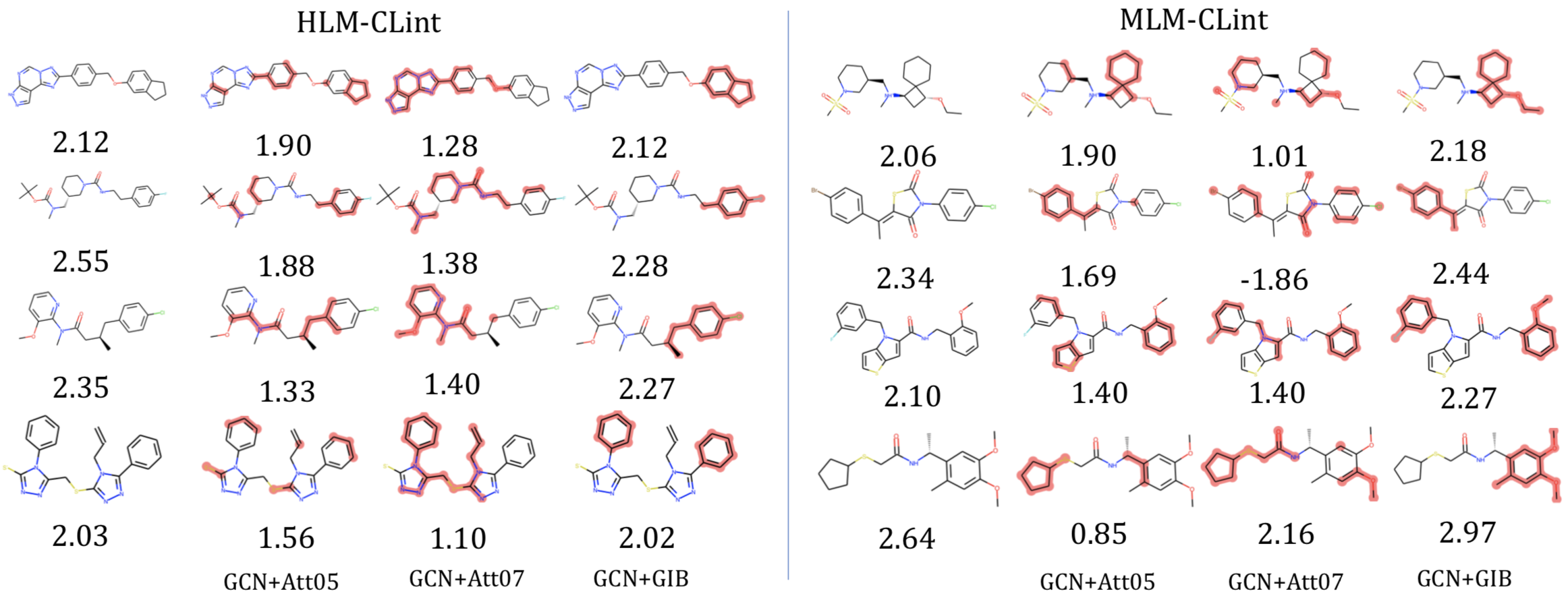}}
\end{center}
\vspace{-0.5cm}
\caption{The molecules with their interpretable subgraphs discovered by different methods. These subgraphs exhibit similar chemical properties compared to the molecules on the left.}
\label{mol-compare}
\end{figure}

\begin{table}[t]
  \centering
  \vspace{-0.2cm}
  \caption{Quantitative results on graph denoising. We report the classification accuracy (Acc), number of real edges over total real edges (Recall) and number of real edges over total edges in subgraphs (Precision) on the test set.}
  
    \begin{tabular}{cccccc}
    \toprule
    Method & {GCN} & {DiffPool} & {GCN+Att05} & {GCN+Att07} & \textbf{GCN+GIB} \\
    \midrule
    Recall & - & - & {0.226$\pm$0.047} & {0.324$\pm$ 0.049} & \textbf{0.493$\pm$ 0.035} \\
    Precision & - & - & {0.638$\pm$ 0.141} & {0.675$\pm$ 0.104} & \textbf{0.692 $\pm$0.061} \\
    Acc   & 0.617 & 0.658 & 0.649 & 0.667 & \textbf{0.684} \\
    \bottomrule
    \end{tabular}%
  \label{tab:edge}%
\vspace{-0.4cm}
\end{table}%

\subsection{Results}

\textbf{Improvement of Graph Classification:} In Table \ref{tab:2}, we comprehensively evaluate the proposed method and baselines on improvement of graph classification. We train GIB on various backbones and aggregate the graph representations only from the subgraphs. We compare the performance of our framework with the mean/sum aggregation and pooling aggregation.  This shows that GIB improves the graph classification by reducing the redundancies in the graph structure.

\begin{wraptable}[8]{r}{0.6\textwidth}
  \centering
  \vspace{-0.7cm}
  \caption{Average number of disconnected substructures per graph selected by different methods.}
    \begin{tabular}{ccccc}
    \toprule
    Method & QED   & DRD2  & HLM   & MLM \\
    \midrule
    GCN+Att05 & 3.38 & 1.94 & 3.11 & 5.16 \\
    GCN+Att07 & 2.04 & 1.76 & 2.75 & 3.00 \\
    \textbf{GCN+GIB}   & \textbf{1.57} & \textbf{1.08} & \textbf{2.29} & \textbf{2.06} \\
    \bottomrule
    \end{tabular}%
  \label{tab:disconnected}%
\end{wraptable}

\textbf{Graph interpretation:} Table \ref{tab:1} shows the quantitative performance of different methods on the graph interpretation task. GIB is able to generate precise graph interpretation (\ibsubgraph), as the substructures found by GIB has the most similar property to the input molecules. Then we derive two variants of our method by deleting $\mathcal{L}_{con}$ and $\mathcal{L}_{MI}$. GIB also outperforms the variants, and thus indicates that every part of our model does contribute to the improvement of performance. In practice, we observe that removing $\mathcal{L}_{con}$ will lead to unstable training process due to the continuous relaxation to the generation of subgraph. In Fig.~\ref{mol-compare}, GIB generates more compact and reasonable interpretation to the property of molecules confirmed by chemical experts. More results are provided in the Appendix. In Table \ref{tab:disconnected}, we compare the average number of disconnected substructures per graph since a compact subgraph should preserve more topological information. GIB generates more compact subgraphs to better interpret the graph property.

\textbf{Graph denoising:} Table \ref{tab:edge} shows the performance of different methods on noisy graph classification. GIB outperforms the baselines on classification accuracy by a large margin due to the superior property of \ibsubgraph. Moreover, GIB is able to better reveal the real structure of permuted graphs in terms of precision and recall rate of true edges.

\section{Conclusion}
In this paper, we have studied a subgraph recognition problem to infer a maximally informative yet compressed subgraph. We define such a  subgraph as IB-subgraph and propose the graph information bottleneck (GIB) framework for effectively discovering an IB-subgraph. We derive  the GIB objective from a mutual information estimator for irregular graph data, which is optimized by a bi-level learning scheme. A connectivity loss is further used to stabilize the learning process. We evaluate our GIB framework in the improvement of graph classification, graph interpretation and graph denoising. Experimental results verify the superior properties of IB-subgraphs.

\bibliography{iclr2021_conference.bib}
\bibliographystyle{plain}

\newpage
\appendix
\section{Appendix}
\subsection{More details about Eq.~\ref{sub-obj} and Eq.~\ref{mi_1}}
\label{appendix_eq9}

Here we provide more details about how to yield Eq.~\ref{sub-obj} and Eq.~\ref{mi_1}.
\begin{equation}
\begin{aligned}
I(Y,\gG_{sub})& = \int p(y,\gG_{sub}) \log{{p(y|\gG_{sub})}} dy \ d\gG_{sub}  - \int p(y,\gG_{sub}) \log{{p(y)}} dy \ d\gG_{sub}  \\
& = \int p(y,\gG_{sub}) \log{{p(y|\gG_{sub})}} dy \ d\gG_{sub}  + \mathrm{H}(Y)   \\
&\geq \int p(y,\gG_{sub}) \log{{q_{\phi_{1}}(y|\gG_{sub})}} dy \ d\gG_{sub} + \mathrm{KL}(p(y|\gG_{sub})|q_{\phi_{1}}(y|\gG_{sub})) \\
&\geq \int p(y,\gG_{sub}) \log{{q_{\phi_{1}}(y|\gG_{sub})}} dy \ d\gG_{sub} \\
&\approx \frac{1}{N} \sum_{i=1}^{N} q_{\phi_{1}}(y_{i}|\gG_{sub_{i}}) \\
&= -\mathcal{L}_{cls}(q_{\phi_{1}}(y|\gG_{sub}),y_{gt}) \\
\end{aligned}
\end{equation}

\subsection{case study}

To understand the bi-level objective to MI minimization in Eq. \ref{bilevel}, we provide a case study in which we optimize the parameters of distribution to reduce the mutual information between two variables. Consider $p(x) = sign(\mathcal{N}(0,1)), p(y|x) = \mathcal{N}(y;x,\sigma^{2})$\footnote{We use the toy dataset from https://github.com/mzgubic/MINE}. The distribution of $Y$ is:

\begin{equation}
\begin{aligned}
p(y) &= \int p(y|x)p(x) dx \\
&=\sum_{i} p(y|x_{i})p(x_{i}) \\
&= p(y|x=1)p(x=1) + p(y|x=-1)p(x=-1) \\
&= 0.5(\mathcal{N}(y;1,\sigma^{2})+\mathcal{N}(y;-1,\sigma^{2}))\\
\end{aligned}
\end{equation}

We optimize the parameter $\sigma^{2}$ to reduce the mutual information between $X$ and $Y$. For each epoch, we sample 20000 data points from each distribution, denoted as $X=\{x_{1},x_{2},\cdots,x_{20000}\}, Y=\{y_{1},y_{2},\cdots,y_{20000}\}$. The inner-step is set to be 150. After the inner optimization ends, the model yields a good mutual information approximator and optimize $\sigma^{2}$ to reduce the mutual information by minimizing $L_{MI}$. We compute the mutual information with the traditional method and compare it with  $L_{MI}$:

\begin{equation}
\begin{aligned}
I(X,Y) &= \int p(x,y) \log{\frac{p(y|x)}{p(y)}} dx dy \\
& \approx \frac{1}{20000} \sum_{i=1}^{20000} \log{\frac{p(y_{i}|x_{i})}{p(y_{i})}}
\end{aligned}
\end{equation}

As is shown in Fig .\ref{mi_est}, the mutual information decreases as $\mathcal{L}_{MI}$ descends. The advantage of such bi-level objective to MI minimization in Eq. \ref{bilevel} is that it only requires samples instead of forms of distribution. Moreover, it needs no tractable prior distribution for variational approximation. The drawback is that it needs additional computation in the inner loop.

\begin{figure}[t]
\begin{center}
\centerline{\includegraphics[width=0.7\columnwidth]{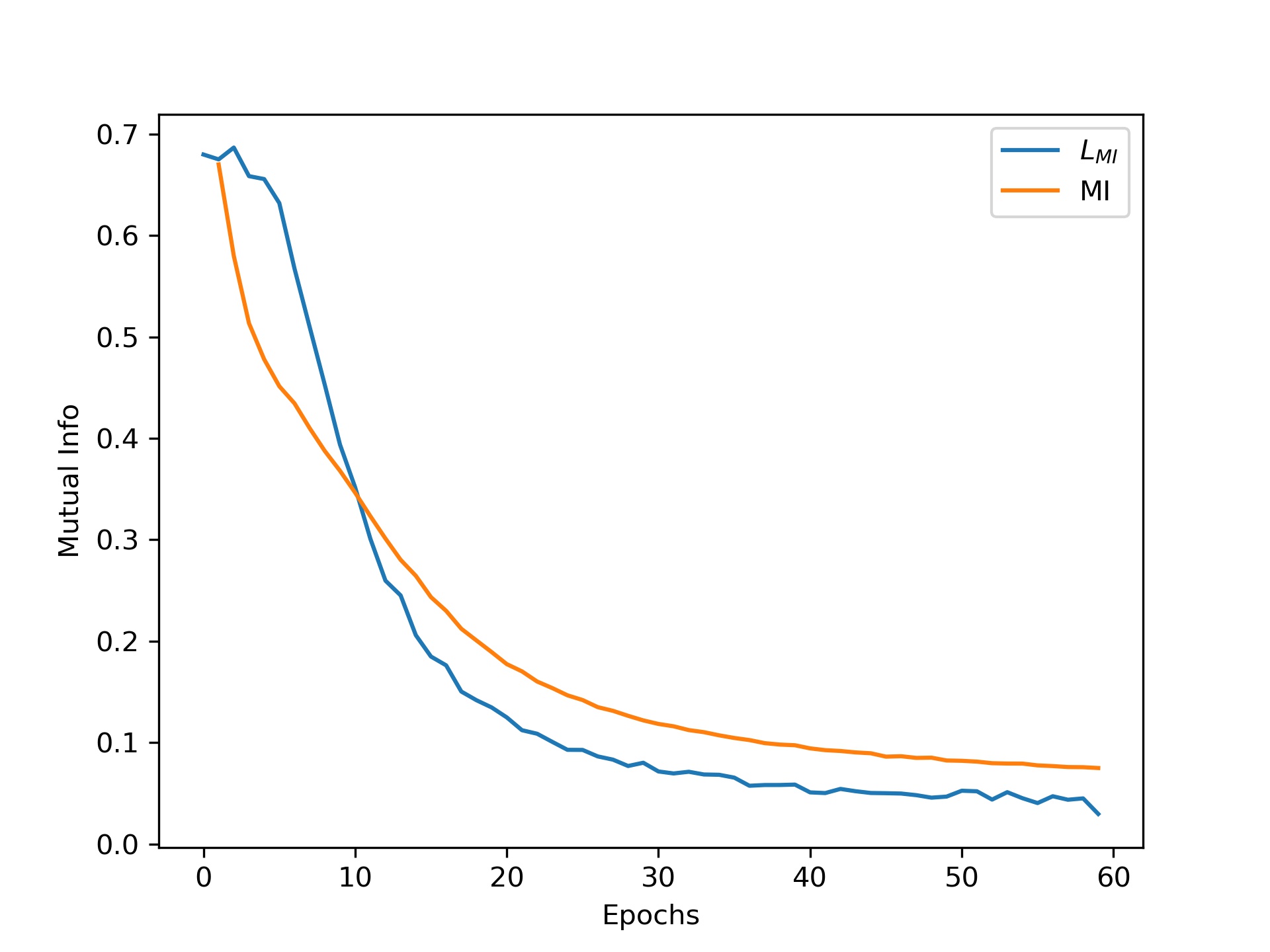}}
\end{center}
\caption{We use the bi-level objective to minimize the mutual information of two distributions. The MI is consistent with the loss as $\mathcal{L}_{MI}$ declines.}
\label{mi_graph}
\end{figure}

\subsection{Algorithm}
    \begin{algorithm}
        \caption{Optimizing the graph information bottleneck.}
        \begin{algorithmic}[1] 
            \Require Graph $\gG=\{\mA,\mX\}$, graph label $Y$, inner-step $T$, outer-step $N$. 
            \Ensure Subgraph $\gG_{sub}$
            \Function {GIB}{$\gG=\{\mA,\mX\}, Y, T, N$}
                \State $\theta \gets \theta^{0},\quad \phi_{1} \gets \phi_{1}^{0}$
                \For{$i=0\to N$}
                    \State $\phi_{2} \gets \phi_{2}^{0}$
                    \For{$t=0 \to T$}
                    
                        \State $\phi_{2}^{t+1} \gets \phi_{2}^{t} + \eta_{1} \nabla_{\phi_{2}^{t}}\mathcal{L}_{\rm MI} $
                    \EndFor
                        
                    \State $\theta^{i+1} \gets \theta^{i} -\eta_{2} \nabla_{\theta^{i}} \mathcal{L}(\theta^{i},\phi_{1}^{i},\phi_{2}^{T})  $
                    \State $\phi_{1}^{i+1} \gets \phi_{1}^{i} -\eta_{2} \nabla_{\phi_{1}^{i}} \mathcal{L}_(\theta^{i},\phi_{1}^{i},\phi_{2}^{T})  $
                \EndFor
                
                \State $\gG_{sub} \gets g(\gG;\theta^{N})$
                \State \Return{$\gG_{sub}$}
            \EndFunction
        \end{algorithmic}
    \end{algorithm}

\subsection{More results on graph interpretation}
\begin{figure}[t]
\begin{center}
\centerline{\includegraphics[width=1.0\columnwidth]{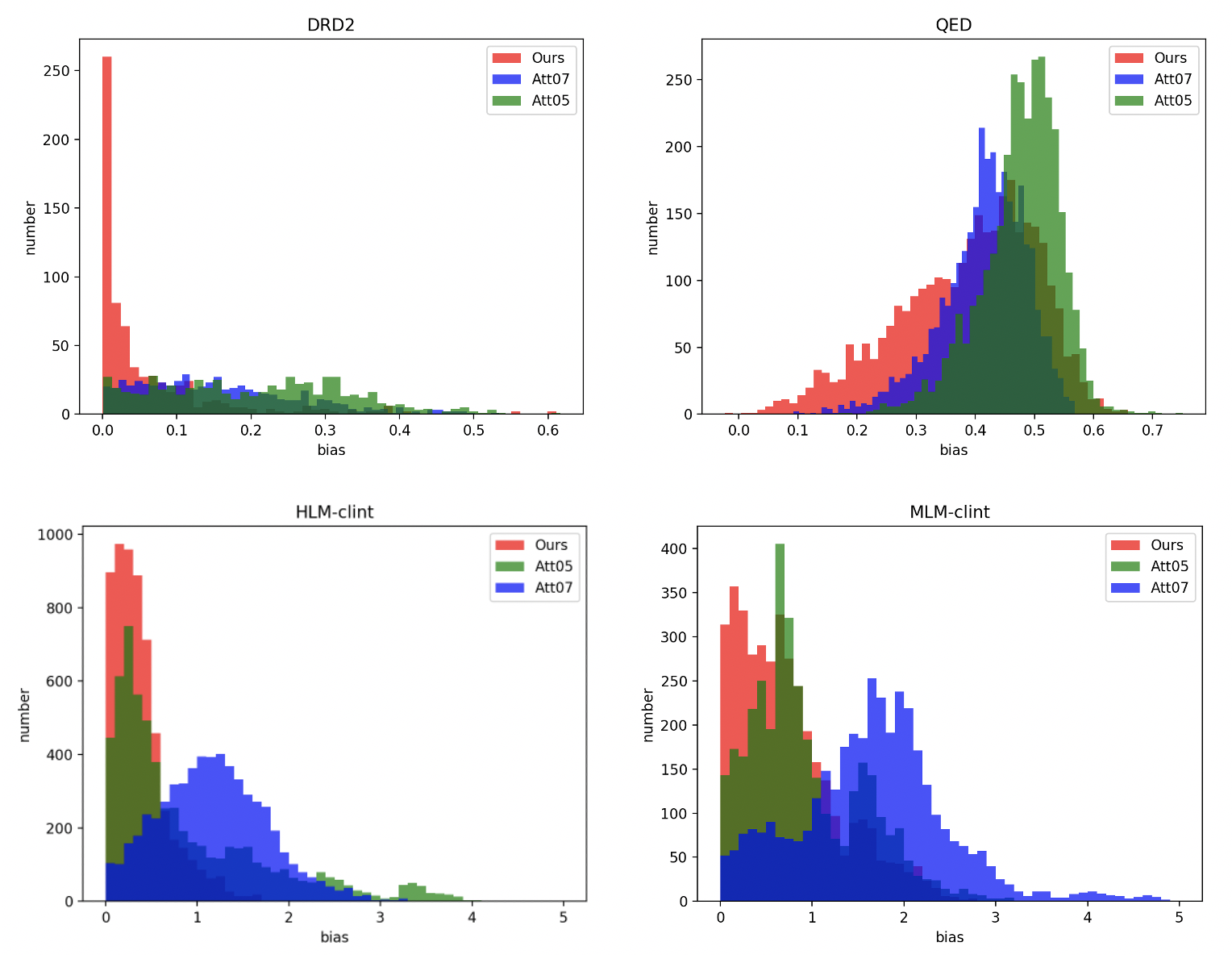}}
\end{center}
\caption{The histgram of absolute bias between the property of graphs and subgraphs.}
\label{hist}
\end{figure}

In Fig.~\ref{hist}, we show the distribution of absolute bias between the property of graphs and subgraphs. GIB is able to generate such subgraphs with more similar properties to the original graphs.

\begin{figure}[t]
\centerline{\includegraphics[width=1.0\columnwidth]{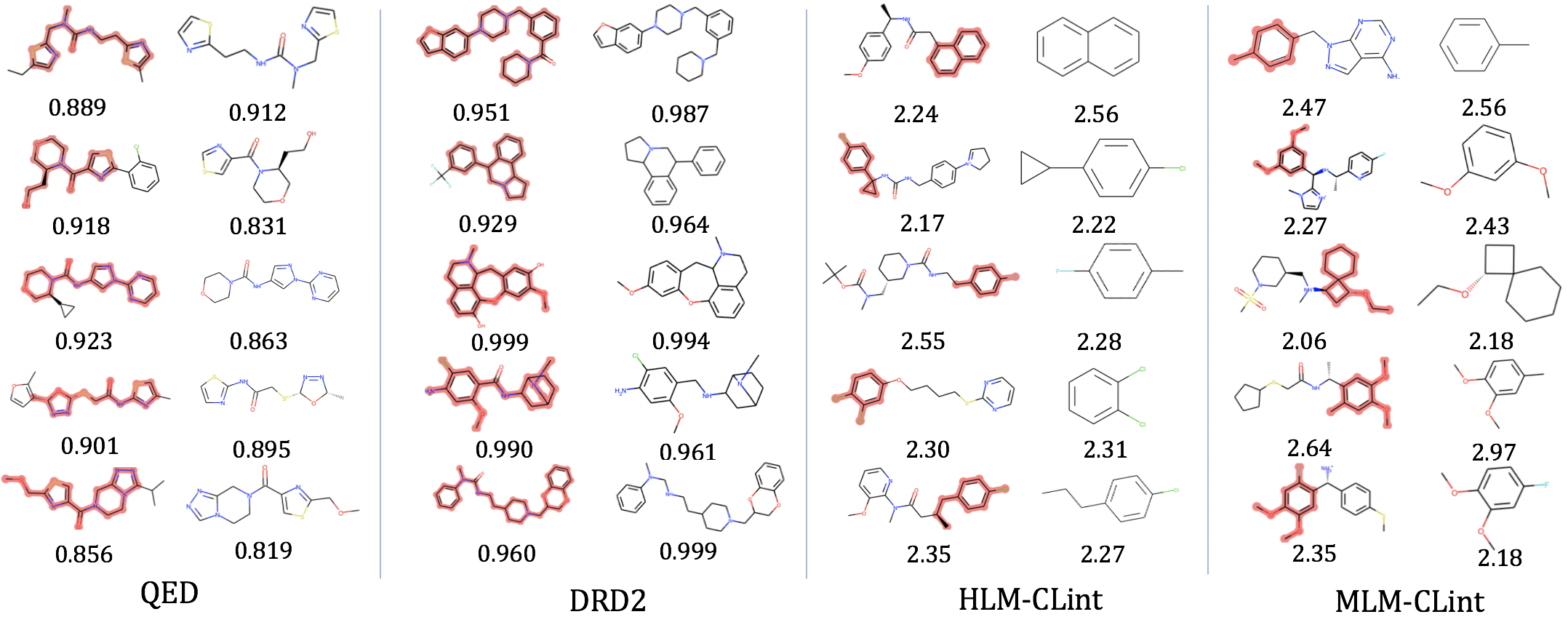}}
\caption{The molecules with its interpretation found by GIB. These subgraphs exhibit similar chemical properties compared to the molecules on the left.}
\label{fig1}
\end{figure}

In Fig.~\ref{fig1}, we provide more results of four properties on graph interpretation. 

\subsection{More results on noisy graph classification}

\begin{figure}[t]
\begin{center}
\centerline{\includegraphics[width=1.0\columnwidth]{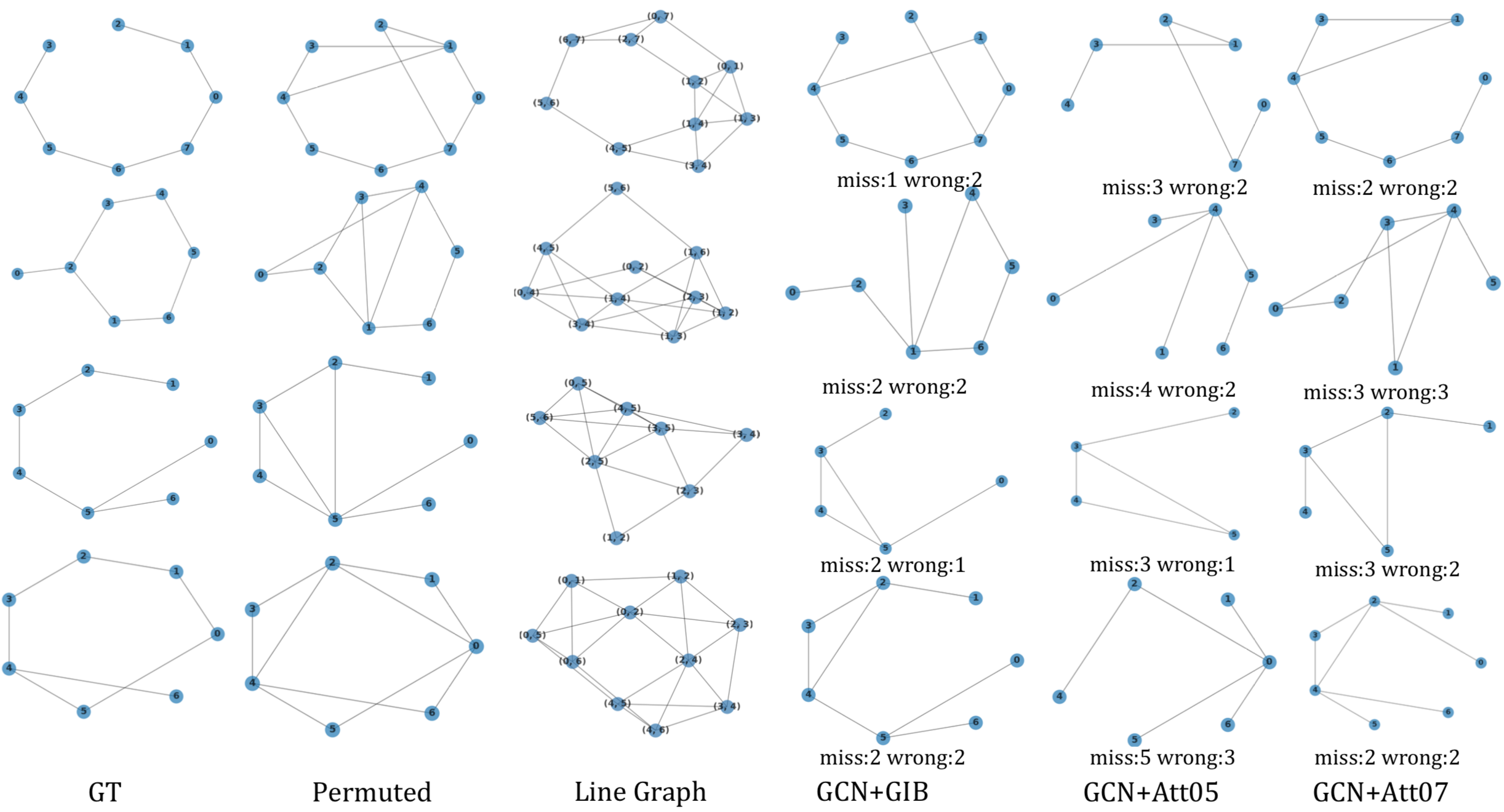}}
\end{center}
\caption{We show the blindly denoising results on permuted graphs. Each method operates on the line-graphs and tries to recover the true topology by removing the redundant edges. 
Columns 4,5,6 shows results obtained by different methods, where ``miss: $m$, wrong: $n$'' means missing $m$ edges and there are $n$ wrong edges in the output graph.
GIB always recognizes more similar structure to the ground truth (not provided in the training process) than other methods. }
\label{newedge}
\end{figure}

We provide qualitative results on noisy graph classification in Fig.~\ref{newedge}.

\end{document}

%% file: math_commands.tex
\usepackage{amsmath,amsfonts,bm}

\def\eqref#1{equation~\ref{#1}}

\def\1{\bm{1}}

\def\mA{{\bm{A}}}

\def\mD{{\bm{D}}}
\def\mE{{\bm{E}}}

\def\mI{{\bm{I}}}

\def\mM{{\bm{M}}}

\def\mS{{\bm{S}}}

\def\mV{{\bm{V}}}
\def\mW{{\bm{W}}}
\def\mX{{\bm{X}}}
\def\mY{{\bm{Y}}}

\DeclareMathAlphabet{\mathsfit}{\encodingdefault}{\sfdefault}{m}{sl}
\SetMathAlphabet{\mathsfit}{bold}{\encodingdefault}{\sfdefault}{bx}{n}

\def\gG{{\mathcal{G}}}

\def\sG{{\mathbb{G}}}

\def\sR{{\mathbb{R}}}

\def\sV{{\mathbb{V}}}

\DeclareMathOperator*{\argmax}{arg\,max}